\documentclass[final]{cvpr}
\usepackage{times}
\usepackage{epsfig}
\usepackage{graphicx}
\usepackage{amsmath}
\usepackage{amssymb}
\usepackage{dsfont}
\usepackage[outline]{contour}

\usepackage{lipsum}
\usepackage{algorithm}
\usepackage{algpseudocode}
\usepackage{subcaption}
\usepackage[table]{xcolor}
\usepackage{diagbox}
\usepackage[pagebackref=true,breaklinks=true,letterpaper=true,colorlinks,bookmarks=false]{hyperref}

\usepackage[pagebackref=true,breaklinks=true,colorlinks,bookmarks=false]{hyperref}

\definecolor{red}{rgb}{1,0,0} 

\begin{document}

\title{Adversarial Imaging Pipelines}

\author{Buu Phan\\
Algolux
\and
Fahim Mannan\\
Algolux
\and
Felix Heide\\
Algolux, Princeton University
}

\maketitle

\begin{abstract}
Adversarial attacks play an essential role in understanding deep neural network predictions and improving their robustness.
Existing attack methods aim to deceive convolutional neural network (CNN)-based classifiers by manipulating RGB images that are fed directly to the classifiers. However, these approaches typically neglect the influence of the camera optics and image processing pipeline (ISP) that produce the network inputs. ISPs transform RAW measurements to RGB images and traditionally are assumed to preserve adversarial patterns. However, these low-level pipelines can, in fact, destroy, introduce or amplify adversarial patterns that can deceive a downstream detector. As a result, optimized patterns can become adversarial for the classifier after being transformed by a certain camera ISP and optic but not for others. In this work, we examine and develop such an attack that deceives a specific camera ISP while leaving others intact, using the same down-stream classifier. We frame camera-specific attacks as a multi-task optimization problem, relying on a differentiable approximation for the ISP itself. We validate the proposed method using recent state-of-the-art automotive hardware ISPs, achieving 92\% fooling rate when attacking a specific ISP. We demonstrate physical optics attacks with 90\% fooling rate for a specific camera lenses.

\end{abstract}

\vspace{-5pt}
\section{Introduction}
Deep neural networks have become a cornerstone method in computer vision \cite{ chen2017deeplab,he2017mask,he2016deep,Isola_2017_CVPR,zhang2016colorful} with diverse applications across fields, including safety-critical perception for self-driving vehicles, medical diagnosis, video security, medical imaging and assistive robotics. Although a wide range of high-stakes applications base their decision making on the output of deep networks, existing deep models have been shown to be susceptible to adversarial attacks on the image that the network ingests. Specifically, existing adversarial attacks perturb the input image with carefully designed patterns to deceive the model while being imperceptible to a human viewer \cite{madry2017towards, papernot2016limitations,poursaeed2018generative,szegedy2013intriguing,nakkiran2019adversarial,tsipras2019robustness}. As such, understanding and exploring adversarial perturbations offer insights into the failure cases of today's models and it allows researchers to develop defense methods and models that are resilient against proposed attacks \cite{borkar2020defending,lu2017safetynet,madry2017towards,papernot2016distillation,xie2019feature}.   
 
 \begin{figure}[t]
\vspace{-15pt}
    \centering
		\includegraphics[width=0.92\linewidth]{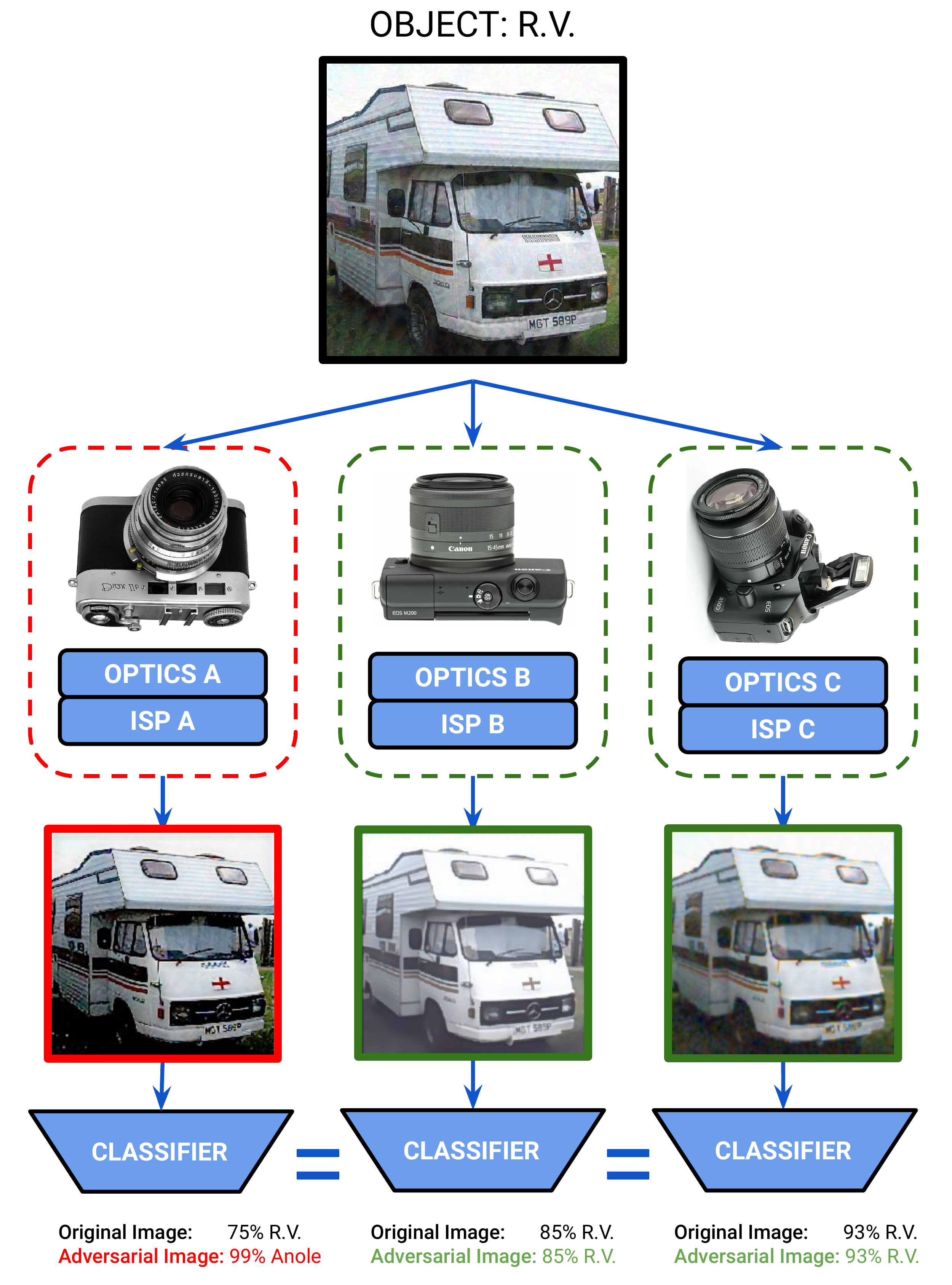}
		\vspace{-10pt}
    \caption{We illustrate and show the camera-specific attack. The image is tampered such that it becomes only adversarial for a specific camera pipeline, even when the three pipelines deploy the same classifier. }
    \label{fig:Teaser}
    \vspace{-20pt}
\end{figure}
Existing adversarial attacks find post-capture adversaries, tampering with the image after capture before it is input to the deep network. Recently, a number of attack methods have been demonstrated in the form of physical objects that are placed in real-world scenes to generate adversarial patterns by capturing images of the physical objects~\cite{athalye2018synthesizing, eykholt2018robust, kurakin2016adversarial}. The most successful methods for computing adversarial perturbations rely on network gradients to form adversarial examples \cite{szegedy2013intriguing, goodfellow2014explaining, kurakin2016adversarial, moosavi2016deepfool, papernot2016limitations, C&W} for each input image, that struggle to transfer to other networks or images \cite{szegedy2013intriguing, transfer_1, transfer_2}. Alternative approaches rely on only the network predictions~\cite{ilyas2018blackbox,black_box_1, single_pixel} and use surrogate networks~\cite{papernot2017practical} or gradient approximations~\cite{obfuscated-gradients}. All of these methods, both physical and synthetic attacks, have in common that they \emph{assume that the camera image processing pipeline (ISP) to preserve the attack pattern}. Although modern image processing pipelines implement complex algorithms, such as tonemapping, sharpening or denoising~\cite{karaimer2016software,kim2012new}, which transform RAW measurements to RGB images on embedded camera processors, the influence of this pipeline is ignored by existing attack methods. Some of the processing blocks in camera image processing pipelines have even been suggested as defenses against existing attacks~\cite{gupta2019ciidefence,liao2018defense}.

In this work, we close this gap between scene-based physical attacks and attacks on post-processed images. Specifically, we propose a novel method that allows us to attack cameras with a specific ISP, while leaving the detections of other cameras intact for the \emph{identical} classifier but a different ISP. As such, the attack mechanism proposed in this work is a \emph{camera-specific attack} that not only targets the deep network but conventional hardware ISPs that traditionally have been not been considered susceptible to adversarial attacks. As a further camera-specific attack, we also attack the optical system of a camera system. The proposed method can incorporate proprietary black-box ISP and complex compound optics, without accurate models, by relying differentiable approximations as gradient oracles. We validate our method using recent automotive hardware ISP processors and automotive optics, where the novel attack achieves a fooling rate of 92\% on RAW images in experimental captures. 

Specifically, we make the following contributions
\begin{itemize}
	\setlength\itemsep{.2em}
	\item We introduce the \emph{first method for finding adversarial attacks that deceives a specific camera ISP} and optics while leaving cameras with other ISPs or optics intact although they employ the same classifier network.
	\item We demonstrate attacks for embedded hardware ISPs that are not differentiable and only available as black-box algorithms. To this end, we learn differentiable approximations of the image processing and sensing pipeline that serves as gradient oracles for our attack.
	\item We analyze and validate the attack on RAW input measurements for state-of-the-art hardware ISPs.
	\item We validate physical attacks of the proposed method on recent automotive camera ISPs and automotive optics. The proposed method achieves more than 90\% success rate.
\end{itemize}

\vspace{-5pt}
\section{Related Work}
Our work considers the problem of adversarial attacks on camera pipelines. We review the relevant literature below.

\vspace{0.2em}\noindent\textbf{Camera Image Processing Pipelines.} Research on high-level vision tasks has often overlooked the existence of the low-level image signal processing (ISP) pipeline in the camera. In practice, the role of these ISPs is critical in a vision system because their ability to recover high quality images from noisy and distorted RAW measurements directly affects the downstream processing modules~\cite{heide2014flexisp,tseng2019hyperparameter}. For display applications, domain-specific image processing methods  ~\cite{gharbi2016deep,GharbiSIGGRAPH2017,chen2018learning,Chen:2017:CoRR,pmlr-v37-xub15,fan2018smoothing,heide2014flexisp} have been successful to tackle low-light, shot-noise and optical aberrations. Unfortunately, these methods are computationally expensive, and, as such, their application is  limited to off-line tasks. In contrast, real-time applications, such as robotics and augmented reality demand real-time processing at more than 30~Hz for double-digit megapixel streams. As a result, integrated system-on-chip ISPs are today employed for robotic vision systems, such as autonomous robots, self-driving vehicles, and drones. For example, the ARM Mali-C71 ASIC ISP is capable of processing 12~megapixel streams at up to 100~Hz with less than one Watt power consumption. However, although hardware ISPs are efficient, these processing pipelines are typically highly optimized proprietary compute units that are not differentiable and their behavior is unknown to the user~\cite{tseng2019hyperparameter}.  In this work, we present the first adversarial attack that targets these hardware processing blocks, which, in contrast to deep neural networks, \emph{traditionally have been assumed to be not susceptible to adversarial perturbations} and instead have been suggested as potential defense units~\cite{gupta2019ciidefence,liao2018defense}.

\vspace{0.2em}\noindent\textbf{Adversarial Attacks.} A large body of work has explored adversarial attacks on deep networks in computer vision. A common formulation describes an attack as an $\ell_p$ norm-ball constrained perturbation that deceives a specific classifier~\cite{madry2017towards}. Depending on the knowledge of the model (\ie weights and architecture) that the adversary has, attacks can be grouped into two settings: white-box and black-box attacks. In the white-box setting, the model specification are known and the adversaries leverage it to synthesize the perturbation. By treating the attack as a solution of an optimization problem, techniques ranges from mixed-integer programming  \cite{tjeng2018evaluating,wong2018provable} to 1\textsuperscript{st}-order gradient method~\cite{goodfellow2014explaining,madry2017towards,moosavi2016deepfool,szegedy2013intriguing} have been proposed. Additionally, by manipulating the optimization objectives and constraints, attacks can reveal interesting properties of the target network, such as sparsity and interpretability~\cite{zhao2020towards,C&W,modas2019sparsefool,xu2018structured,tao2018attacks}. In the black-box setting, adversaries can only query the input-output pairs, and, hence, the target model is more difficult to be deceived. Nevertheless, existing approaches have shown that adversaries can successfully approximate the gradients and apply the white-box method. This is achieved by approximating the target network function~\cite{papernot2017practical,obfuscated-gradients} (transfer methods) or by numerical estimation (score methods) \cite{huang2019black,tu2019autozoom,chen2020frank,li2019nattack,chen2017zoo}. In this work, we propose a transfer approach that approximates non-differentiable camera pipelines, including the camera optics  and ISP, with differentiable proxy functions. 

Going beyond synthetically generated adversarial examples, researchers have shown to be able to recreate them in the wild by placing adversarial patterns on physical objects. Kurakin~{\it{et al.}}~\cite{kurakin2016adversarial} demonstrate such a physical attack by printing the digital adversarial image on paper and capturing it with a camera, assuming that the acquisition and capture pipeline itself is not susceptible to the adversarial pattern. Athalye~{\it{et al.}}~\cite{athalye2018synthesizing} propose an attack which optimizes the perturbation under different image augmentations, a direction further explored by a line of work~\cite{eykholt2018robust,jan2019connecting,song2018physical,duan2020adversarial,chen2018shapeshifter} to achieve higher attack ratios. All of these existing methods have in common that they assume that the scene light transport and acquisition preserve the adversarial patterns, including the optics, sensors and ISP in the camera as non-susceptible image transforms. As a direct result, \emph{existing physical attacks have failed to achieve the high fooling rates of synthetic attacks}~\cite{kurakin2016adversarial}. Our work fills this gap and, as a result, shows that it is possible to achieve high fooling rates when including the acquisition and processing operations in adversarial attacks. Building on this insight, we realize attacks of individual camera types by exploiting slight differences in their acquisition and image processing pipeline.



\begin{figure*}[t]
\vspace{-0.5cm}
    \centering
    \includegraphics[width=1.0\textwidth]{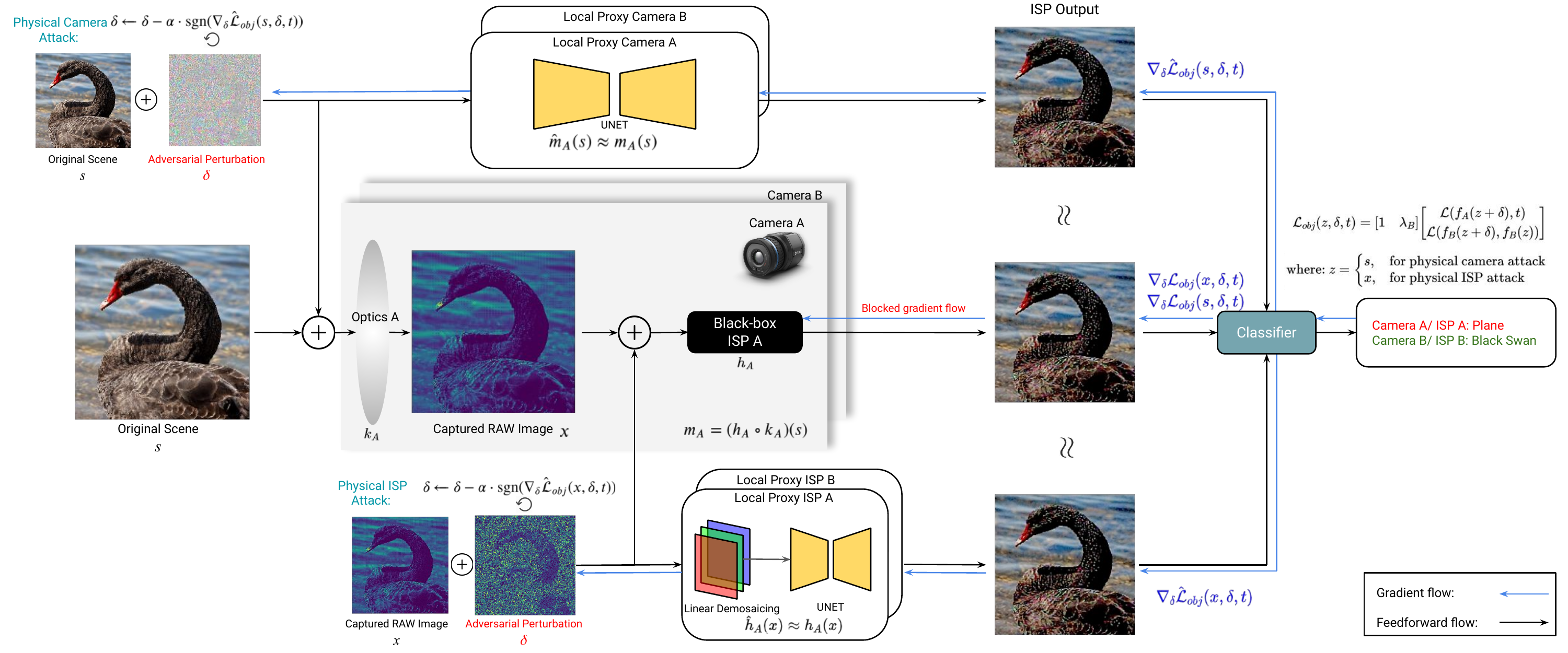}
    \vspace{-0.87cm}
    \caption{ Overview of the proposed targeted camera attack. We perturb either the display scene (physical camera attack) or the captured RAW image (physical ISP attack), whose label is ``black swan'',  such that they are misclassified into ``plane'' by pipeline A but not by pipeline B. To find such an attack, we solve an optimization problem, using the estimated gradients from the proxies approximation of the black-box, non-differentiable imaging modules. The objective function is a weighted sum of two cross-entropy losses, where the first term encourages the attack to fool pipeline A and the second term prevents it from changing the original prediction probability of pipeline B.} 
    \label{fig:ProxyNetworks}
    \vspace{-0.5cm}
\end{figure*}

\vspace{-5pt}
\section{Background}
In this section, we review the differentiable proxy framework from Tseng et al.~\cite{tseng2019hyperparameter} and the projected gradient descent $\ell_p$ norm-bounded adversarial attack~\cite{madry2017towards}, and we introduce relevant notation for the following sections. 

\subsection{Differentiable Proxy ISPs}
A given non-differentiable hardware ISP is approximated by a differentiable proxy function, which implements a mapping from RAW input data to post-ISP images via a convolutional neural network (CNN). We note that this framework can also be extended to include the compound optics in the pipeline (see the Supplementary Document).

\vspace{0.2em}\noindent\textbf{Proxy ISP Model.} We denote $h: \mathbb{R}^d \xrightarrow[]{} \mathbb{R}^{d\times3}$ as a black-box ISP function that maps a RAW image $x\in \mathbb{R}^d$ to an RGB image, where $d$ is the RAW image dimension (\eg $1920\times1200$). The proxy ISP function $\Tilde{h}_\theta: \mathbb{R}^d \xrightarrow[]{} \mathbb{R}^{d\times3}$ depends on $\theta$ as learnable parameters (\ie CNN weights) also maps a RAW image to a post-ISP image. As a departure from Tseng et al.~\cite{tseng2019hyperparameter}, we found that bilinear demosaicing as a first layer in this proxy module improves training stability and accuracy. This demosaicing layer is differentiable. The demosaiced RGB image is fed into a U-Net~\cite{ronneberger2015u}, which is trained to approximate the output of the hardware ISP. 

\vspace{0.2em}\noindent\textbf{Proxy Training.} Given a set of RAW captures: $X=\{ x_1, x_2,..., x_N \}$ where each $x_i \in \mathbb{R}^{d}$, we train the proxy function $\Tilde{h}_\theta$ by minimizing the $\ell_1$ reconstruction loss. 

\subsection{Projected Gradient Adversarial Attacks}\label{adv_attack_bg}
Let us denote a probabilistic classifier that maps an input $x \in \mathbb{R}^d$ to a categorical distribution vector as $f : \mathbb{R}^d \xrightarrow[]{} \mathbb{R}^K$, where $d$ is the input dimension and $K$ is the number of classes. 
We define a decision function $c(x)$, which assigns a label to $x$ as: $c(x) = \underset{k=1,2,...,K}{\arg\max} f^{k}(x)$. 

\vspace{0.2em}\noindent\textbf{$\ell_p$ norm-bounded attack.} For an input $x$, an additive perturbation $\delta \in \mathcal{B}_d(p;\epsilon)$ is adversarial when  $c(x+\delta) = t$, where $t$ is a target label and $\mathcal{B}_d(p;\epsilon) =\{r\in R^d:\|r\|_p < \epsilon\}$ is an $\ell_p$ norm-ball with radius $\epsilon$\footnote{As an image, $(x+\delta)$ needs to stay within the valid range (\eg [0,255] for RGB images), which can be achieved by the clipping operation. We implicitly assume this condition throughout this paper without stating it.}. We will use $\ell_\infty$ throughout this paper. To create such a perturbation, we solve the following constrained optimization problem
\begin{equation}
\begin{aligned}
& \underset{\|\delta\|_\infty \leq \epsilon}{\text{minimize}}
& & \mathcal{L}(f(x+\delta), t),
\label{lp_opt}
\end{aligned}
\end{equation}
where $\mathcal{L}$ is the cross-entropy loss. 

\vspace{0.2em}\noindent\textbf{Projected Gradient Descent (PGD).}\label{Iterative_attack} In the case of $\ell_\infty$, we can  solve (\ref{lp_opt}) by first randomly initializing $\delta \in \mathcal{B}_d(\infty;\epsilon)$ and iteratively perform the following PGD update
\begin{equation}
\delta \xleftarrow[]{} \delta - \alpha \cdot \text{sgn} (\nabla_\delta \mathcal{L}(f(x + \delta),t)).
\end{equation}
where $\alpha$ is the step size, which can depend on the current ordinal iteration  number. This attack can be denoted as a ``targeted'' attack. The ``untargeted'' variant, where we aim to fool the model independently of the target class, can be formulated in a similar fashion by maximizing the loss in (\ref{lp_opt}) and set the target label to the original prediction.

\vspace{-5pt}
\section{Camera Pipeline Adversarial Attack}

In the following, we consider an camera pipeline consisting of a black-box, non-differentiable ISP followed by a downstream RGB image classifier. A direct RAW attack on such a pipeline involves manipulating the captured RAW image. For a physical camera attack, our pipeline also includes the optical system that captures an adversarial scene. In this section we only explain the direct RAW attack without any loss of generality.

Next, we describe two types of attacks on these pipelines and the method to generate them. The first type of attack, referred to as \emph{untargeted camera attack}, aims to craft an adversarial RAW perturbation to the pipeline, without considering its transferability to the other pipelines. The second type, referred to as \emph{targeted camera attack}, generates a perturbation that deceives a specific pipeline while leaving the other intact, even when the same classifier is deployed. Figure~\ref{fig:ProxyNetworks} provides an overview of the proposed targeted camera attack and corresponding proxy functions.

We define a black-box ISP function as $h : \mathbb{R}^{d} \xrightarrow[]{} \mathbb{R}^{d \times 3}$, a trained proxy function that approximates $h$ as $\tilde{h}_\theta: \mathbb{R}^{d} \xrightarrow[]{} \mathbb{R}^{d \times 3}$  and an RGB image classifier as $g : \mathbb{R}^{d \times 3} \xrightarrow[]{} \mathbb{R}^K$. Given a RAW image $x \in \mathbb{R}^d$, we define the camera pipelines using the original ISP and proxy ISP separately as: $f(x)=(g\circ h)(x)$ and $\tilde{f}(x)=(g\circ \tilde{h})(x)$. Similar to Sec.\ref{adv_attack_bg}, $c(x)$ and $\tilde{c}(x)$ are the corresponding decision functions. Before describing the two camera attacks, we next introduce the a local proxy function, which is a modification of Tseng et al.~\cite{tseng2019hyperparameter}'s model that is essential to the success of the proposed attack.

\begin{figure}[t]
\vspace*{-3ex}
\begin{algorithm}[H]
\caption{Local Proxy Training} 
\label{attack1algo}
\begin{algorithmic}[1]
\small
\renewcommand{\algorithmicrequire}{\textbf{Input:}}
 \renewcommand{\algorithmicensure}{\textbf{Output:}}
\Require $h$; $\tilde{h}$; $g$; Number of augmented images $M$; number of attack iterations $n$; a list of targeted images $S$; a predefined bound $\epsilon$; update step size $\alpha$.
\Ensure A local proxy function $\hat{h}$
\State $\hat{S} = S$
\State $\tilde{f}=(g\circ \tilde{h})$
\ForAll{$x_i \in S$ }:
\For{$m \xleftarrow[]{}1...M$}: 
\State $\hat{\epsilon} \thicksim \text{uniform}(\alpha, \epsilon + \alpha)$
\State $\delta \xleftarrow[]{} \text{PGD}(x_i,\tilde{f}, n, \hat{\epsilon}; \alpha)$  \Comment{\parbox[t]{.42\linewidth}}{perform n-steps PGD update, target random class} 
\State $\hat{S} = \hat{S} \cup \{x_i+\delta, h(x_i+\delta)\}$
\EndFor
\EndFor 

\State $\hat{h} \xleftarrow[]{} \text{TRAIN}(\tilde{h},\hat{S})$  \Comment{Train the local proxy  $\hat{h}$ from $\hat{S}$ and $\tilde{h}$ } \\
\Return $\hat{h}$ 
\end{algorithmic}
\end{algorithm}
\vspace{-30pt}
\end{figure}

\subsection{Local Gradient Proxies}\label{type1}
 In our experiments, we found that, despite $\tilde{h}(x)$ being perceptually similar to $h(x)$, performing the PGD-update based on the estimated gradient from $\tilde{h}$ does not result in high success rate in many cases, especially for the targeted camera attack. To this end, we propose using  a local proxy model as an alternative gradient-oracle, which is trained by fine-tuning the existing proxy model $\tilde{h}$ with a set of target images and Jacobian augmentation \cite{papernot2017practical}. We find that such a local proxy model effectively improves the success rate for both untargeted and targeted camera attacks.

 Specifically, given an image set $S$ that we wish to attack, we create $M$ different Jacobian-augmented pairs: $\{(x_i+\delta_i), h(x_i+\delta_i)\}$ for each $x_i\in S$, where $\delta_i \in \mathcal{B}_d(p;\hat{\epsilon})$ is the adversarial perturbation on the proxy pipeline $\tilde{f}$ and the bounded radius $\hat{\epsilon}$ is uniformly sampled within $[\alpha, \epsilon + \alpha]$, where $\alpha$ is the PGD update step size. The local proxy model $\hat{h}$ is obtained by finetuning $\tilde{h}$ with the newly augmented training set $\hat{S}$. This method is formalized in Algorithm~\ref{attack1algo}.
 

\subsection{Untargeted Camera Attack}\label{type1}

\begin{figure}[t]
    \centering
		\includegraphics[width=1.0\linewidth]{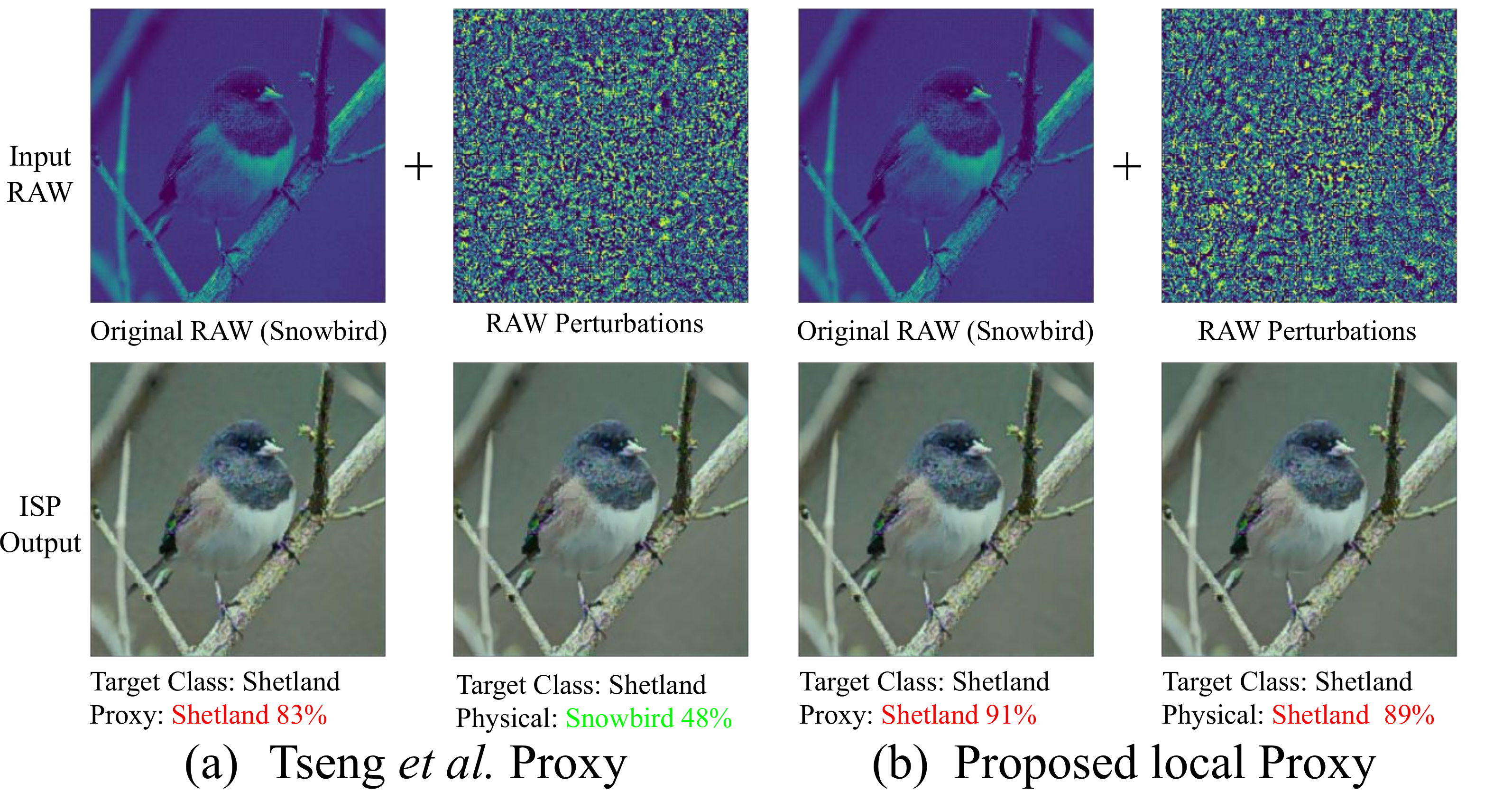}
		\vspace{-22pt}
    \caption{Local Proxies: Example that demonstrates the proxy from Tseng \textit{et al.}~\cite{tseng2019hyperparameter} fails to approximate the real ISP adequately for adversarial attack. The proposed local proxy attack successfully causes the physical pipeline to misclassify the image into the target class ``Shetland'', while Tseng \textit{et al.}'s proxy fails.
    }
    \label{fig:localISP_example}
    \vspace{-15pt}
\end{figure}

For this attack type, we aim to generate an adversarial perturbation $\delta \in \mathcal{B}_d(\infty;\epsilon)$ to a RAW image $x$ such that: $c(x+\delta) = t$ independent of the camera pipeline. We replace the black-box ISP $h$ with its local proxy function $\hat{h}$ and generate adversarial perturbations $\delta$ from the PGD update on $\hat{f} = g \circ \hat{h}$, that is:\begin{equation}
\delta \xleftarrow[]{} \delta - \alpha \cdot \text{sgn} (\nabla_\delta \mathcal{L}(\hat{f}(x + \delta),t)) 
\vspace{-5pt}
\end{equation}
We found that, despite both $\tilde{h}(x),\hat{h}(x)$  being perceptually similar to $h(x)$, estimated gradient using $\hat{h}$ consistently yields a higher success rate than  $\tilde{h}$ (refer to the supplementary document for quantitative comparisons). We illustrate this in Figure~\ref{fig:localISP_example}, showing that being trained with different perturbations enables $\hat{h}$ to provide accurate gradients for the attack to transfer well to $h$.

\subsection{Targeted Camera Attack}

For this attack type, we aim to craft a perturbation that deceives a specific camera pipeline $h$, while leaving the classifications of other camera pipelines intact, even when all the pipelines deploy the same classifier $g$. Let $h_i$, for $i\in \{1,2,...,T\}$, be one of the ISPs that we do not want to attack, its associated camera pipeline and decision function are $f_i(x)=(g\circ h_i)(x)$ and $c_i(x)$. We assume that $g$ is transferable across different ISPs, \ie accuracy higher than $70\%$ for each ISP in the Imagenet dataset. Ideally, an adversarial perturbation $\delta \in \mathcal{B}_d(\infty;\epsilon)$ to an image $x$ with label $y$, should satisfy: $c(x+\delta) = t$ and $f_i(x+\delta) =f_i(x)$, given that $c(x)=y$ and every $c_i(x)=y$. Such a perturbation can be found as a solution of the following optimization problem
\begin{equation}
\begin{aligned}
& \underset{||\delta||_p \leq \epsilon}{\text{minimize}}
& &  \mathcal{L}(f(x+\delta), t) \\
& \text{s.t}
& & f_i(x+\delta) = f_i(x) , \forall i \in {1,...,T} .\\
\end{aligned} \label{attack2_org}
\end{equation}
The problem from (\ref{attack2_org}) is a challenging nonlinear-equality constrained problem that may have only have a feasible solution with large cross-entropy loss.

\vspace{0.2em}\noindent\textbf{Soft-Constrained Objective}\label{relaxed_problem}  
When $h$ and $h_i$ are known and differentiable, we can relax (\ref{attack2_org}) using soft-constraints and applying the PGD update on $\delta$ to jointly minimize the objective function and the distance between $f_i(x+\delta) $ and $ f_i(x)$
\begin{equation}
\begin{aligned}
& \underset{||\delta||_p \leq \epsilon}{\text{minimize}}
& & \mathcal{L}_{obj}(x,\delta,t) ,\\
\end{aligned} \label{attack2_relax}
\end{equation}
where $\mathcal{L}_{obj}(x,\delta,t) = \mathcal{L}(f(x+\delta),t) + \sum_{i=1}^{T} \lambda_i\mathcal{L}(f_i(x+\delta),f_i(x))$. The second term measures the cross-entropy loss between $f_i(x+\delta) $ and $ f_i(x)$ and each $\lambda_i$ is set to 1 in our experiment. We note that minimizing the cross-entropy loss in this case is equivalent to minimizing the KL divergence between the two categorical distributions as
\begin{equation}
\begin{aligned}
 \mathcal{L}(\!f_i(x\!+\!\delta), f_i(x))&{=} \mathbb{D}_{KL}(\!{f}_i(x\!+\!\delta)|| f_i(x)\!) \!-\! \mathbb{H}(\!f_i(x))  \\
&{=} \mathbb{D}_{KL}(\!{f}_i(x\!+\!\delta)|| f_i(x))\!+\! \text{const} .
\end{aligned} \label{KL}
\end{equation} 

\vspace{0.2em}\noindent\textbf{Objective Function with Local Proxy ISP.} Since $h$ and $h_i$ can be non-differentiable, we optimize $\delta$ on the new objective function, which replace $h$ and $h_i$ with their corresponding local proxy $\hat{h}$  and $\hat{h}_i$, that is
\begin{equation}
\hat{\mathcal{L}}_{obj}(x,\delta,t){=} \mathcal{L}(\hat{f}(x\!+\!\delta),t){+}\!\!\sum_{i=1}^{T}\! \lambda_i\mathcal{L}(\hat{f}_i(x\!+\!\delta),f_i(x)).\label{new_obj_hat}
\end{equation} 
The ISP-specific perturbation is found by performing the PGD update on $\hat{\mathcal{L}}_{obj}$, that is
\begin{equation}
\delta \xleftarrow[]{} \delta - \alpha \cdot \text{sgn} (\nabla_\delta \hat{\mathcal{L}}_{obj}(x,\delta,t)) .
\end{equation}
We note that replacing $h$ and $h_i$ with $\tilde{h}$ and $\tilde{h}_i$ does not give a high success rate, even for large $\epsilon$. This is because the gradient estimation quality from $\tilde{h},\tilde{h_i}$ is not accurate enough for satisfying several constraints. Finally, while the proposed objective (\ref{new_obj_hat}) only minimizes the KL divergence between $\hat{f}_i(x+\delta)$ and $ f_i(x)$, training the local proxy model has indirectly minimized the distance between $\hat{f}_i(x+\delta)$ and $f_i(x+\delta)$ around the perturbation radius $\epsilon$. We formulate this method in Algorithm~\ref{attack2algo}.

\begin{figure}[t]
\vspace{-25pt}
\begin{algorithm}[H]
\caption{Targeted Camera Adversarial Perturbation} 
\label{attack2algo}
\begin{algorithmic}[1]
\small
\renewcommand{\algorithmicrequire}{\textbf{Input:}}
 \renewcommand{\algorithmicensure}{\textbf{Output:}}
\Require Targeted ISP $h$; Untargeted ISPs $\{h_1,h_2,...,h_T\}$; Pretrained local proxies: $\hat{h},\{\hat{h}_1,\hat{h}_2,...,\hat{h}_T\}$; RGB classifier $g$; number of attack iterations $n$; targeted image $x$; targeted class $t$; perturbation bound $\epsilon$.
\Ensure  adversarial image $x' \in \mathbb{R}^d$
\State \textit{/ / Construct the proxy pipelines:}
\State $\hat{f} = (g \circ \hat{h}); \hat{f}_i = (g \circ \hat{h}_i) $
\State \textit{/ / Construct the objective function:}
\State $\hat{\mathcal{L}}_{obj}(x,\delta,t) = \mathcal{L}(\hat{f}(x+\delta),t) + \sum_{i=1}^{T} \lambda_i\mathcal{L}(\hat{f}_i(x+\delta),f_i(x))$
\State \textit{/ / Attack the targeted image:}
\State $\delta \thicksim \text{uniform}(-\epsilon, \epsilon)$
\For{$k \xleftarrow[]{}1...n$}: 
\State $\delta \xleftarrow[]{} \text{clip}(x + \delta) - x$ \Comment{\parbox[t]{.5\linewidth}}{Clip $\delta$ to the valid range}
\State $\delta \xleftarrow[]{} \delta - \alpha \cdot \text{sgn} (\nabla_\delta \hat{\mathcal{L}}_{obj}(x,\delta,t))$ 
\EndFor
\State $x' = \text{clip}(x + \delta)$ \Comment{\parbox[t]{.5\linewidth}}{Clip $x+\delta$ to the valid range}
\\
\Return $x'$
\end{algorithmic}
\end{algorithm}
\vspace{-25pt}
\end{figure}

\vspace{-5pt}
\section{Assessment} \label{exp}
We validate our methods using hardware ISPs and optical assemblies for direct RAW and physical camera attacks.

\subsection{Validation Experiments}
\vspace{-0.2em}\noindent\textbf{Dataset.} For all the experiments, we use a subset of 1,000 ImageNet validation images ~\cite{imagenet}.

\vspace{0.2em}\noindent\textbf{Image Processing Pipelines.} We evaluate our method for the black-box/non-differentiable hardware ARM Mali C71 and Movidius Myriad 2 ISPs. In addition to the two hardware ISPs, we also jointly evaluate with two differentiable ISPs. The first one only performs bilinear demosaicing, and will be referred to as the Demosaicing ISP. The second one performs bilinear demosaicing operation followed by bilateral filtering ~\cite{paris2009bilateral}, and referred to as Bilateral Filter ISP. All ISPs are described in detail in the Supplementary Document.

\vspace{0.2em}\noindent\textbf{Optics.} We use a Fujinon CF12.5HA-1 lens with 54\textdegree{} field of view as the default lens for our experiments. As this compound optics is a proprietary design, we evaluate the proposed attacks on a Cooke triplet optimized for image quality using Zemax Hammer optimization and fabricated using PMMA. Details are described in the Supplementary Document.

\vspace{0.2em}\noindent\textbf{Classifier.} We use a large Resnet-101 ~\cite{imagenet} classifier, which achieves $76.4\%$ Top-1 accuracy. Since each ISP has a different set of parameters (such as white-balance coefficients, color-correction matrix, etc.), we prevent the domain-shift problem by finetuning the pretrained Resnet-101 model on a set of ISP output images.

\vspace{0.2em}\noindent\textbf{Evaluation Metrics.} We evaluate success rate, transfer rate and targeted success rate as metrics in our evaluation. Success rate measures whether an attack for a given camera pipeline is able to change that pipeline's prediction to the target class. Transfer rate measures whether an adversarial RAW is misclassified by other pipelines. Targeted success rate measures if an attack pattern changes the targeted pipeline's prediction to the target class while leaving other camera pipelines unaffected (class prediction does not change and the confidence difference between the original and adversarial RAW is below 0.15). 

\vspace{-5pt}\subsection{Physical Setup}
\begin{figure}[t]
\vspace{-12pt}
    \centering
		\includegraphics[width=1.0\linewidth]{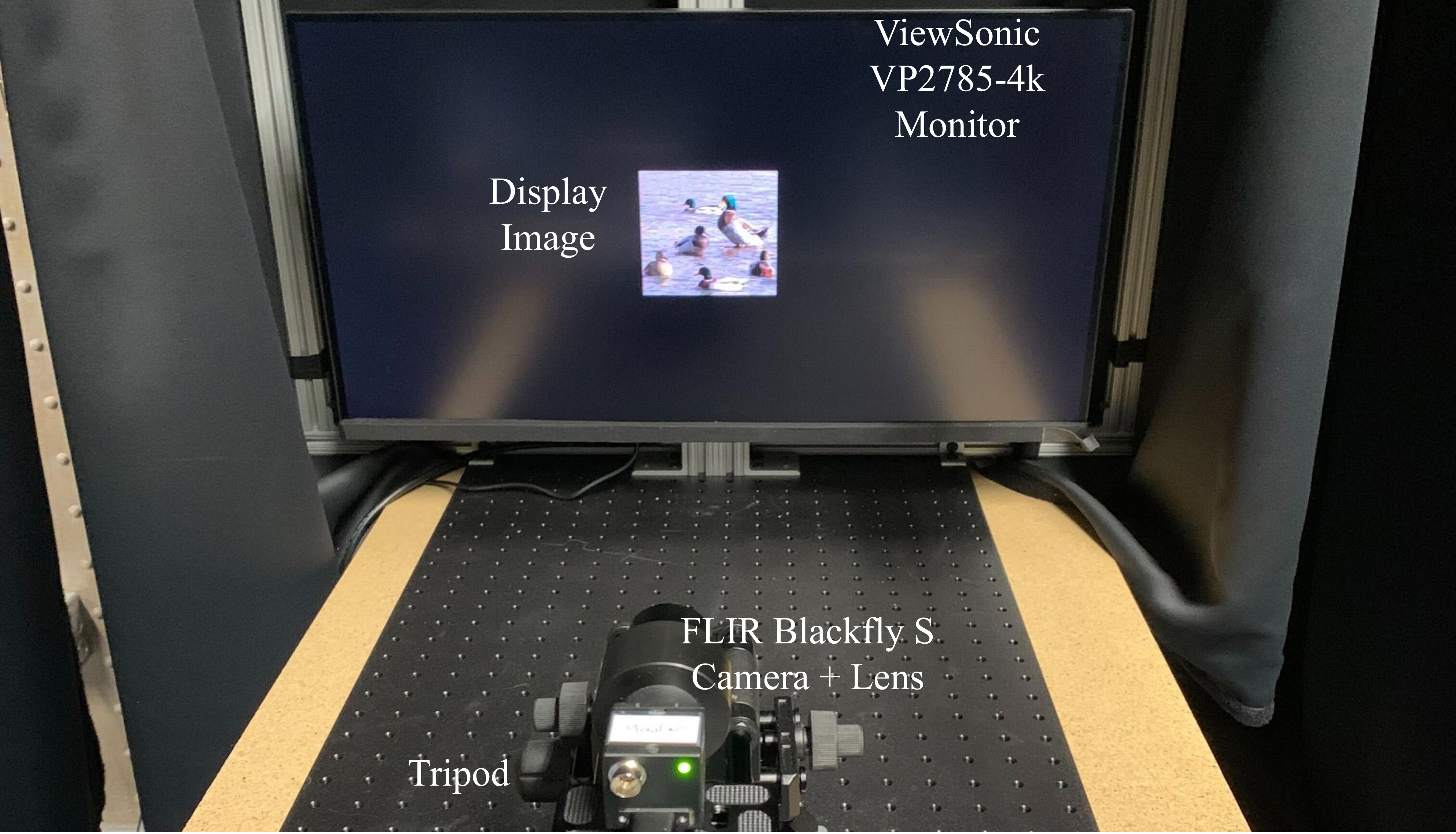}
		\vspace{-18pt}
    \caption{Setup for Evaluation of Camera-Specific Attacks. We employ a monitor placed in front of the target camera system, which is attacked by the proposed method. The proposed setup allows us to evaluate attacks on specific cameras, including their ISPs and camera optics using physical captures.}
    \label{fig:Physical Setup}
    \vspace{-10pt}
\end{figure}
To validate the proposed method in a physical setup, we display the attacked images on the ViewSonic VP2785-4k monitor, as shown in Figure~\ref{fig:Physical Setup}. This setup allows us to collect large-scale evaluation statistics in a physical setup, departing from sparse validation examples presented in existing works with RGB printouts~\cite{kurakin2016adversarial,athalye2018synthesizing}. We capture images using a FLIR Blackfly S camera employing a Sony IMX249 sensor. The camera is positioned on a tripod and mounted such that the optical axis aligns with the center of the monitor. The camera and monitor are connected to a computer, which is used to jointly display and capture thousands of validation images. Each lens-assembly is focused at infinity with the screen beyond the hyperfocal distance. The captured RAW image acquired by the sensor in this setup is fed to the ISPs and then resized to the resolution of $224\times224$ before going through the Resnet-101 classifier.

\vspace{-5pt}\subsection{Physical ISP Attack}
In this attack setting, we acquire the RAW images by projecting the images onto the screen, using the FLIR camera. These RAW images are then fed to a hardware ISP and the adversarial perturbation is added directly to the RAW image. We use  $\epsilon=2000$ to reliably deceive a RAW image while keeping the perturbation imperceptible. For each image, we target a random class and use a total of 30 iterations, with step size $\alpha=50$.

\begin{table}[t]
    \begin{subtable}{1.0\linewidth}
      \centering\resizebox{1.0\textwidth}{!}{
\renewcommand{\arraystretch}{1.1}%
\begin{tabular}{|c|cccc|}
\hline
\backslashbox{\textbf{Targeted ISP}}{\textbf{Deployed ISP}} & Movidius Myriad 2    & ARM Mali C71    & Bilateral Filter ISP  & Demosaicing ISP\\ \hline
Movidius Myriad 2 &\textbf{93.5\%} &28.4\% $\|$ 50.6\%   &31.3 \% $\|$  53.6\%   &39.7\% $\|$ 59.4\%               \\
ARM Mali C71  &34.3\%$\|$46.1\% &\textbf{94.4\%}    &14.5\%$\|$30.3\% &18.7\%$\|$44.3\%               \\
Bilateral Filter ISP   &29.8\%$\|$44.8\% &18.9\%$\|$32.4\% & \textbf{97.3\%}  &94.3\%$\|$97.3\%             \\
Demosaicing ISP &25.2\%$\|$45.3\% &23.3\%$\|$35.1\% &40.4\%$\|$66.6\%&  \textbf{98.2\%}   \\ \hline
\end{tabular}}
\vspace{-6pt}
        \caption{Untargeted Physical ISP Attack}
    \end{subtable}%
    
\smallskip
    \begin{subtable}{1.0\linewidth}
      \centering
        \resizebox{1.0\textwidth}{!}{
\renewcommand{\arraystretch}{1.1}%
\begin{tabular}{|c|cccc|}
\hline
\backslashbox{\textbf{Targeted ISP}}{\textbf{Deployed ISP}} & Movidius Myriad 2    & ARM Mali C71    & Bilateral Filter ISP  & Demosaicing ISP\\ \hline 

Movidius Myriad 2 &\textbf{92.2\%} &4.3\% $\|$ 5.4\%   &0.0 \% $\|$  0.0\%   &0.0 \% $\|$  0.0\%               \\
ARM Mali C71    &4.8\%$\|$7.1\% &\textbf{93.8\%}    &0.0 \% $\|$  0.0\% &0.0 \% $\|$  0.0\%               \\
Bilateral Filter ISP  &4.7\%$\|$6.9\% & 4.5\%$\|$5.6\% & \textbf{97.3\%}  &0.0 \% $\|$  0.0\%             \\
Demosaicing ISP &4.8\%$\|$7.0\% &4.2\%$\|$5.1\% &0.0 \% $\|$  0.0\%&  \textbf{98.2\%}   \\ \hline
\end{tabular}}
\vspace{-6pt}
\caption{Targeted Physical ISP Attack}
    \end{subtable}
    \vspace{-8pt}
    \caption[Caption for LOF]{Success and transfer rate  for the proposed (a) untargeted and (b) targeted physical ISP attack. Each row shows the attack success rate on the targeted ISP (diagonal cells) and transfer rate to other ISPs (non-diagonal cells\footnotemark). The proposed targeted method significantly reduces the transfer rate across different ISPs. }
\label{Physical ISP}
 \vspace{-16pt}
\end{table}
\footnotetext{The first number of the non-diagonal cell is the transfer rate. The second number measures the percentage of images whose confidence for the adversarial image significantly differs from the adversarial-free one (if their confidence difference is greater than 0.15). }

\begin{figure*}[h]
    \centering
		\vspace{-0.8cm}
    \includegraphics[width=0.99\linewidth]{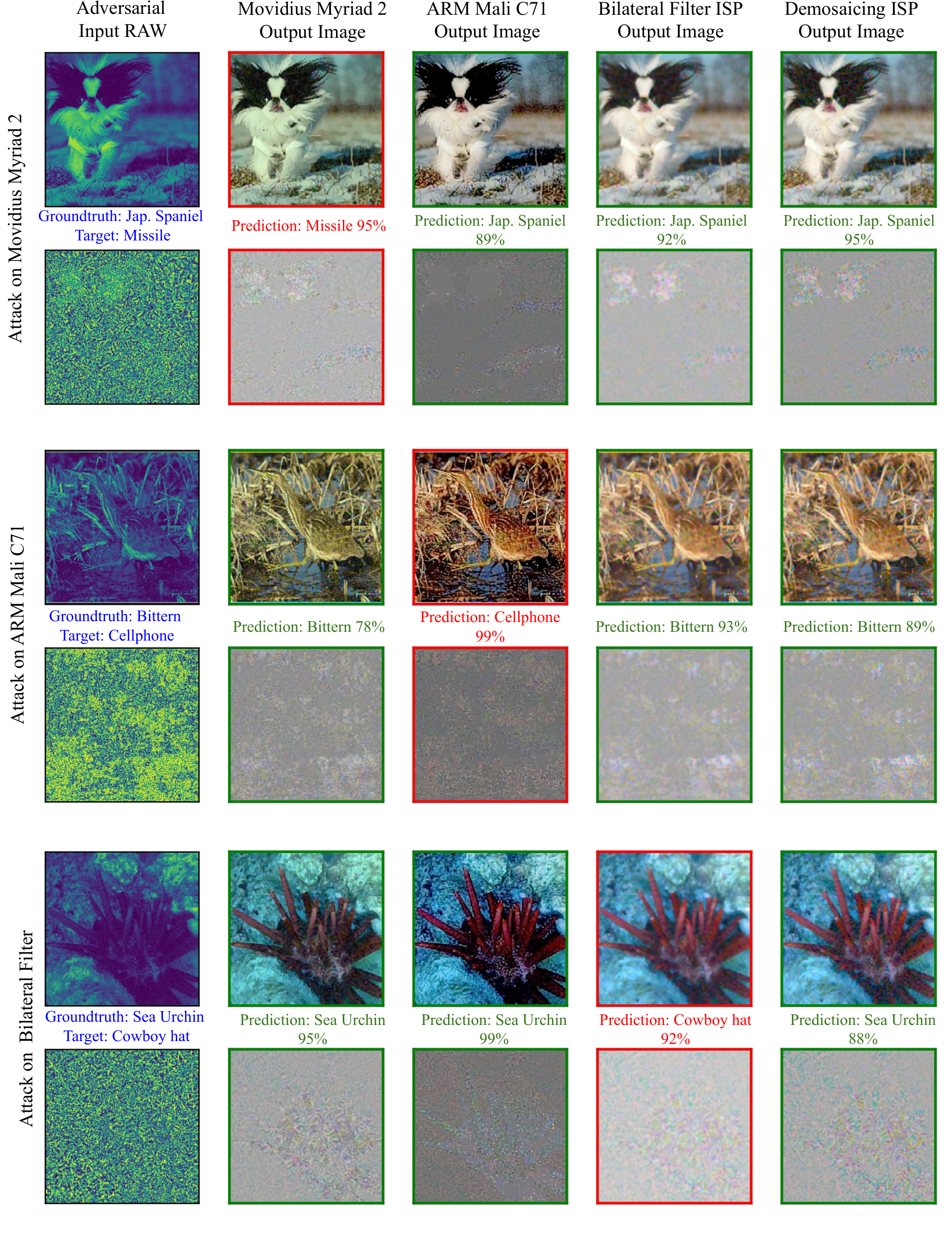}
    \vspace*{-0.8cm}
    \caption{Visualization of the adversarial images and perturbations for the targeted ISP attack. Each pair of rows (top to bottom) shows the attack on the Movidius Myriad 2, ARM Mali C71 and Bilateral Filter ISP respectively. In each targeted ISP attack, we show in the first column the adversarial RAW (top) and perturbations (bottom). The next four columns show the associated  RGB images and perturbations from the ISPs. The RGB perturbation is visualized by subtracting the ISP output of adversarial RAW to that of the unattacked output.}
    \label{fig:ISP attacks}
    \vspace{-0.5cm}
\end{figure*}

\vspace{0.2em}\noindent\textbf{Untargeted Camera Attack.} We measure the transferability of the untargeted camera attack described in Sec.~\ref{type1} in Table~\ref{Physical ISP}a. We observe that the attacks on one ISP are more transferable to certain ISPs, \textit{e}.\textit{g}., attacks on the ARM Mali C71 ISP are more transferable to the Movidius Myriad 2 than the two differentiable ones. Also, attacks on the Bilateral Filter ISP and Demosaicing ISP are likely to be transferable to each other, but not to the hardware ISPs.

\vspace{0.2em}\noindent\textbf{Targeted Camera Attack.} We employ the proposed targeted attack Algorithm~\ref{attack2algo} to craft an attack that only comes into effect when fed into a specific pipeline. We show the result in Table~\ref{Physical ISP}b, where our method significantly reduces the transfer rate across different ISPs. For each targeted black-box ISP attack, it reduces the transfer rate of the Bilateral Filter ISP and Demosaicing ISP to $0.0\%$, and the transfer rate to the other black-box ISP is reduced to below $8.0\%$. 

Figure~\ref{fig:ISP attacks} shows the adversarial RAW images, perturbations (targeting on different ISPs) and their associated ISP outputs. Interestingly, despite having the same adversarial RAW image as input, each ISP produces distinct RGB perturbations. For example, in the attack on Movidius Myriad 2, unlike other ISPs, the ARM Mali C71 suppresses the perturbation around the top left black regions. Also, while the output RGB perturbations seem to contain similar macro structures, only the one from the targeted ISP becomes adversarial to the classifier, while others pose no threat at all. Since the untargeted RGB perturbations do not change the prediction, it means that they are considered as noise by some hidden projections in the classifier. As such, the perturbations are specifically tailored to a specific ISP. In general, for each targeted ISP, our method is able to deceive the target pipeline with more than $87\%$ success rate\footnote{See  Supplementary Document for results per ISP.}. 


\begin{table}[t]
\vspace{-15pt}


      \centering
        \resizebox{0.45\textwidth}{!}{
\renewcommand{\arraystretch}{1.1}%
\begin{tabular}{|c|cc|}
\hline
\backslashbox{\textbf{Targeted Optics}}{\textbf{Deployed Optics}}    & Fujinon CF12.5HA-1    & Cooke Triplet    \\ \hline

Fujinon CF12.5HA-1 &\textbf{90.7\%} &4.5\%$\|$7.9\%               \\
Cooke Triplet    &5.2\%$\|$8.1\% &\textbf{91.5\%}        \\
\hline
\end{tabular}}
\vspace{-6pt}
\caption{Success and transfer rate for the targeted physical optics attack. Refer to Table~\ref{Physical ISP} for table notation.}
\vspace{-12pt}
    
\label{Physical Optics}
 \vspace{-5pt}
\end{table}

\begin{figure}[t]
\vspace{-20pt}
    \centering
		\includegraphics[width=.995\linewidth]{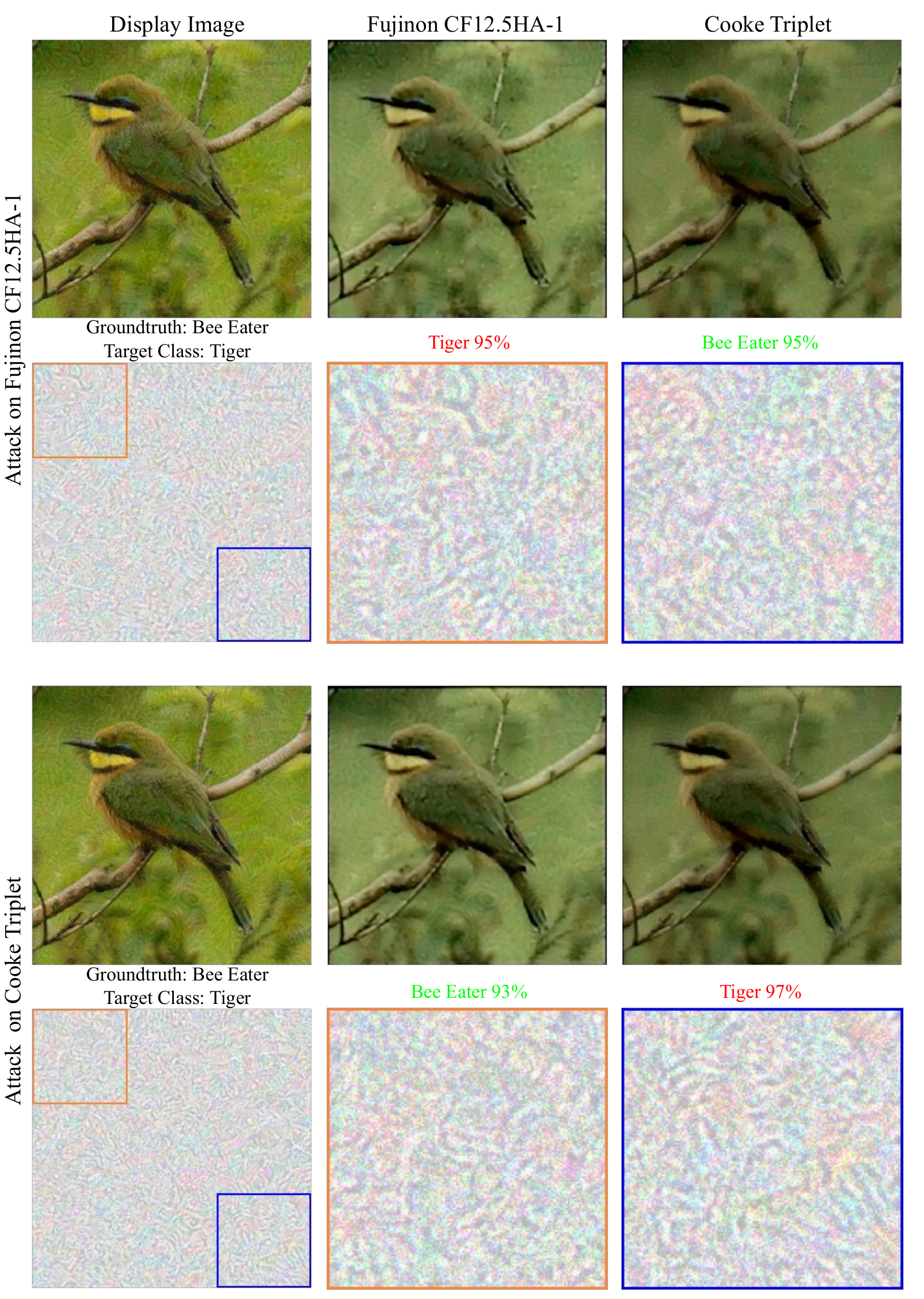}
		\vspace{-22pt}
    \caption{Visualization of the targeted optics attack on the Fujinon CF12.5HA-1  and Cooke Triplet optics. For each attack, we show the displayed adversarial and  post-processed images (top row). In the bottom row, we visualize (from left to right) the additive perturbations on the display image and its zoomed in $150\times150$ top-left and bottom right region. }
    \label{fig:Optics Attack Visualization}
    \vspace{-18pt}
\end{figure}
\vspace{-5pt}\subsection{Attacking Camera Optics}
We extend the proposed method to target a compound optical module instead of a hardware ISP. The proxy function now models the entire transformation from the displayed image to optics, sensor, and ISP processing that results in the final RGB image that is fed to the image classifier. In these experiments, all the pipelines deploy \emph{identical} ARM Mali C71 ISP, which allows us to assess adversarial pattern that targets only one optical system but not another. For each attacked image, we use  $\epsilon=0.08$, target a random class and use a total of 30 iterations, with the step size $\alpha=0.005$. We note that the value of $\epsilon$ is larger since we need to compensate for the attenuation loss during the acquisition process. We apply the same Algorithm~\ref{attack2algo} for the targeted optics attack and show its success and transfer rate \footnote{See Supplementary Document for the untargeted optics attack.} in  Table~\ref{Physical Optics}. The proposed method is able to achieve a high success rate of 90\% while keeping the transfer rate lesser than 10\%. We visualize the attacks in Figure~\ref{fig:Optics Attack Visualization}. We find that in both attacks, the perturbations show distinctive frequency-dependent patterns. We interpret this attack as one that efficiently exploits the frequency bands specific to the optical transfer functions of the employed optics.



\vspace{-5pt}
\section{Conclusion}
In this work, we introduce the \emph{first method for finding adversarial attacks that deceives a specific camera ISP and optics} while leaving cameras with other ISPs or optics intact although they employ the \emph{same classifier network}. Departing from existing adversarial attacks, that assume camera pipelines to preserve adversarial perturbations, we propose an optimization method that employs a local proxy network, making it possible to attack embedded hardware ISPs that are not differentiable and only available as black-box algorithms. We validate the method experimentally on recent automotive camera ISPs and optics, achieving more than 90\% targeted success rate for both ISP and optics attacks. 

Building on the proposed methods, we envision not only research on defense mechanism to improve future image processing and camera optics, but the method also suggests end-to-end multimodel sensor design as a potential avenue to design systems resilient against adversarial attacks. 
\vspace{-5pt}
{\small
\bibliographystyle{ieee_fullname}
\bibliography{egbib}
}

\end{document}